\documentclass[10pt,final,twocolumn]{IEEEtran}
\long\def\comment#1{}

\UseRawInputEncoding

\usepackage{textcomp}
\usepackage{cite}
\usepackage{graphicx}
\usepackage{subfig}
\usepackage{subfloat}
\usepackage{psfrag}
\usepackage{url}
\usepackage{stfloats}
\usepackage{amsmath}
\usepackage{amssymb,amsfonts}
\usepackage{amsthm}
\usepackage{float}
\usepackage[usenames]{color}
\usepackage{algorithm,algorithmic}
\usepackage{booktabs}
\usepackage{longtable}
\usepackage{tabularx}
\usepackage{graphicx}
\usepackage{multirow}
\usepackage{multicol}

\begin{document}

\setlength{\arraycolsep}{0.3em}

\title{Composition and Application of Current Advanced Driving Assistance System: A Review
\thanks{}}

\author{Xinran~Li,
        Kuo-Yi~Lin,~\IEEEmembership{Member,~IEEE,}
        Min~Meng,
        Xiuxian~Li,~\IEEEmembership{Member,~IEEE,}
        Li~Li,~\IEEEmembership{Member,~IEEE,}
        Yiguang~Hong,~\IEEEmembership{Fellow,~IEEE,}
        and~Jie~Chen,~\IEEEmembership{Fellow,~IEEE}

\thanks{This research was supported by National Key R$\&$D Program of China, No. 2018YFE0105000, 2018YFB1305304 and the Shanghai Municipal Commission of Science and Technology No. 19511132100, 19511132101. \em{(Corresponding author: Li Li.)}}
\thanks{All the authors are with the Department of Control Science and Engineering, College of Electronics and Information Engineering, and the Shanghai Institute of Intelligent Science and Technology, Tongji University, Shanghai 201804, China (e-mail: yingli98@tongji.edu.cn, 19603@tongji.edu.cn, mengmin@tongji.edu.cn, xli@tongji.edu.cn, lili@tongji.edu.cn, yghong@tongji.edu.cn, chenjie206@tongji.edu.cn).}
\thanks{M. Meng and Xiuxian Li are also with the Institute for Advanced Study, Tongji University, Shanghai, 200092, China.}
}

\maketitle

\setcounter{equation}{0}
\setcounter{figure}{0}
\setcounter{table}{0}

\begin{abstract}

Due to the growing awareness of driving safety and the development of sophisticated technologies, advanced driving assistance system (ADAS) has been equipped in more and more vehicles with higher accuracy and lower price. The latest progress in this field has called for a review to sum up the conventional knowledge of ADAS, the state-of-the-art researches, and novel applications in real-world. With the help of this kind of review, newcomers in this field can get basic knowledge easier and other researchers may be inspired with potential future development possibility.

This paper makes a general introduction about ADAS by analyzing its hardware support and computation algorithms. Different types of perception sensors are introduced from their interior feature classifications, installation positions, supporting ADAS functions, and pros and cons. The comparisons between different sensors are concluded and illustrated from their inherent characters and specific usages serving for each ADAS function. The current algorithms for ADAS functions are also collected and briefly presented in this paper from both traditional methods and novel ideas. Additionally, discussions about the definition of ADAS from different institutes are reviewed in this paper, and future approaches about ADAS in China are introduced in particular.

\end{abstract}


\section{Introduction}\label{intro}

The development of automated driving (AD) is nowadays to meet the driving assistant demands in real-world situations. The architecture of automated driving can be categorized as environment perception, behavior planning and motion control \cite{zhu2017overview}, according to the autonomous execution process. Based on the implementation of automated driving and the responsibility of driving behaviors, the Society of Automotive Engineers (SAE)  released a taxonomy and classified automated driving from $level$ $0$ (fully human control) to $level$ $5$ (fully automated driving) \cite{TaxonomyAD}. In reference to the classification of SAE, ADAS merely reaches up to $level 2$, which assumes duties such as detecting the surrounding of vehicles, warning drivers for emergency and carrying out one or more simple control functions including for example speed control and adaptation, emergency brake execution and so forth. Currently, a few research institutes realize autonomous driving as far as the  functions in $level$ $3$ and most of commercial vehicles only support self-driving up to $level$ $2$ to make a full assistance in human driving \cite{kuutti2020survey}. In other words, the existing implementation of automated driving mainly reaches up to the class of ADAS. For this reason, it is worthwhile to make a detailed introduction about ADAS and summarize current applied ADAS functions and implementations.

ADAS, as one of the most significant systems embedded in commercial vehicles, focuses on reducing human-related errors and avoiding the amount and the severity of potential traffic accidents \cite{eskandarian2019research}. According to the summary of 2019-2020 MWL-Associated Open Safety Recommendations \cite{national2020most}, eliminated distractions, implementing a comprehensive strategy to reduce speeding-related crashes, increasing implementation of collision avoidance systems in all new highway vehicles and reducing fatigue-related accidents are all in the top $10$ of most wanted list of transportation safety improvements. In fact, all the above requirements can all be components of ADAS.

ADAS was first introduced in the commercial vehicle market with the feature of ABS  (Antilock Braking System) in late 1980s \cite{gordon2015automated}, which works as a passive safety assistance for vehicle insurance. These kinds of ADAS functions were only provided as luxury features for upscale brands in 20 century and in early 2000. However, ADAS has developed rapidly since 2000, thanks to the related automated driving competitions and challenges such as the DARPA Urban Challenge (2007) \cite{buehler2009darpa}, the VisLab Intercontinental Autonomous Challenge (2010) \cite{broggi2010development}, the Grand Cooperative Driving Challenge (2011) \cite{geiger2012team} and so on. It is now a complex and well-developed system composing of both hardware architecture for perception and software design for post processing \cite{galvani2019history}. It improves the driving comfort by helping or reshaping driving behaviors and avoids traffic crashes to ensure driving safety \cite{ziebinski2016survey}. Moreover, with the technology development and the reduced price of hardware sensors, ADAS has generally become a standard and indispensable equipment embedded in current vehicles to provide exterior information \cite{lu2005technical}, warn drivers in time and even exert simple control, and offer additional support for the vehicles.

\begin{table*}[]
\centering
\caption{Overview of Current Reviews}
\label{overview}
\resizebox{\textwidth}{!}{%
\begin{tabular}{@{}|l|c|c|l|@{}}
\toprule
\textbf{Paper}                                                                     & \multicolumn{1}{l|}{\textbf{Year}} & \multicolumn{1}{l|}{\textbf{AD/ADAS}} & \textbf{Contribution}                                                                                                                                                                                                                                        \\ \midrule
A Survey of Autonomous Driving: Common Practices and Emerging Technologies\cite{yurtsever2020survey}      & 2020                               & AD                                    & \begin{tabular}[c]{@{}l@{}}a. the system design and architecture of automated driving   \\ b. related sensors and hardware features \\ c. detailed introduction of localization, mapping, perception, planning and decision making\end{tabular}              \\ \midrule
Self-driving Cars: A survey\cite{badue2020self}                                                           & 2020                               & AD                                    & \begin{tabular}[c]{@{}l@{}}a. perception introduction with related sensor usage and algorithms\\ b. decision making introduction\\ c. architecture of their intelligent vehicle design\end{tabular}                                                          \\ \midrule
A Review of Sensor Technologies for Perception in Automated Driving\cite{marti2019review}                 & 2019                               & AD                                    & \begin{tabular}[c]{@{}l@{}}a. inherent characters and corresponding technologies of perception sensors\\ b. the application of sensors in automated driving\\ c. history of automated driving perception\end{tabular}                                        \\ \midrule
Autonomous Vehicle Perception: The Technology of Today and Tomorrow\cite{van2018autonomous}               & 2018                               & AD                                    & \begin{tabular}[c]{@{}l@{}}a. history and background information about automated driving \\ b. sensors served for automated driving and the potential future improvements about sensor applications\\ c. localization approaches and algorithms\end{tabular} \\ \midrule
Review of Advanced Driver Assistance Systems (ADAS)\cite{ziebinski2017review}                             & 2017                               & ADAS                                  & \begin{tabular}[c]{@{}l@{}}a. the usage introduction of ADAS\\ b. inherent characters of related sensors\end{tabular}                                                                                                                                        \\ \midrule
Perception, Planning, Control, and Coordination for Autonomous Vehicles\cite{pendleton2017perception}     & 2017                               & AD                                    & \begin{tabular}[c]{@{}l@{}}a. inherent features of each sensor and their related software algorithms\\ b. introduction of planning algorithms\\ c. implementation of control\end{tabular}                                                                         \\ \midrule
Overview of Environment Perception for Intelligent Vehicles\cite{zhu2017overview}                         & 2017                               & AD                                    & \begin{tabular}[c]{@{}l@{}}a. briefly introduction of each sensor\\ b. detection technology introduction based on the software implement process\end{tabular}                                                                                                \\ \midrule
A Survey of ADAS Technologies for the Future Perspective of Sensor Fusion\cite{ziebinski2016survey}       & 2016                               & ADAS                                  & \begin{tabular}[c]{@{}l@{}}a. the usage introduction of ADAS\\ b. inherent characters of related sensors\end{tabular}                                                                                                                                        \\ \midrule
A Survey of Motion Planning and Control Techniques for Self-driving Urban Vehicles\cite{paden2016survey}  & 2016                               & AD                                    & \begin{tabular}[c]{@{}l@{}}a. decision making\\ b. planning and control\end{tabular}                                                                                                                                                                         \\ \midrule
A Review of Motion Planning Techniques for Automated Vehicles\cite{gonzalez2015review}                    & 2016                               & AD                                    & a. motion planning                                                                                                                                                                                                                                           \\ \midrule
Automated Driving and Autonomous Functions on Road Vehicles\cite{gordon2015automated}                     & 2015                               & AD                                    & \begin{tabular}[c]{@{}l@{}}a. evolution of control implementation\\ b. perception sections for highly automated driving\end{tabular}                                                                                                                              \\ \midrule
Three Decades of Driver Assistance Systems: Review and Future Perspectives\cite{bengler2014three}         & 2014                               & ADAS                                  & \begin{tabular}[c]{@{}l@{}}a. history and background information about driver assistance\\ b. current state of related technology and researches\\ c. stimuli for future developments\end{tabular}                                                           \\ \midrule
Advanced Driver Assistance Systems - Past, Present and Future\cite{shaout2011advanced}                    & 2011                               & ADAS                                  & a. embedded unit design for each ADAS functions                                                                                                                                                                                                              \\ \bottomrule
\end{tabular}%
}
\end{table*}

The related research for ADAS indeed has nowadays raised much attention in both institutes and development departments in many companies. The amount of published papers about the research and implementation of ADAS functions has been on the rise in the past five years. In 2019 alone, there were more than $500$ related papers published on the Web of Science \cite{hafeez2020insights}. Academic groups from Carnegie Mellon University, Stanford University, Cornell University, University of Pennsylvania and Peking University all built up their own intelligent vehicles for the researches in this field \cite{zhu2017overview}. Companies, like Audi, BWM, Volvo, also invested a lot for the improvement and deployment of ADAS.

To conclude current development of ADAS and provide possible research direction, it is necessary to make a summary about ADAS with the state-of-the-art researches and to present details in ADAS implementation and hardware constructions. However, to our best knowledge, the current reviews and surveys are usually merely focusing on the developments or latest news about autonomous vehicles. ADAS, the fully realized element of self-driving car, is just briefly introduced as a small section in the current reviews about autonomous vehicles. As seen in Table I, we make a summary of the current related papers. Most of the latest reviews about intelligent vehicles are autonomous driving related, which commonly focus on the architecture of the whole system \cite{yurtsever2020survey} or mainly pay attention to the high level of automated driving, such as decision-making \cite{badue2020self, paden2016survey} and control \cite{gordon2015automated, pendleton2017perception, gonzalez2015review}. Even in the reviews written specifically for the perception function of self-driving, the sensors are proposed for their inherent features but presented in isolation with the total construction of vehicles \cite{marti2019review, zhu2017overview}. Papers for ADAS specific functions usually introduce each function by talking about the history development \cite{bengler2014three} and current usage in real-world traffic situations \cite{ziebinski2016survey, ziebinski2017review} rather than their implementation and detailed sensor usage.

By analyzing the above published papers and references therein for basic sensor characters and based on the hardware constructions posted on vehicle manufacture websites, we introduce sensor information from their own features, the installation characters on vehicles and their special design suitable for ADAS functions. This summary is specifically useful for the newcomers to the ADAS field to obtain the basic knowledge about the hardware selection and constructions. We also collect the latest researches about the implementation of algorithms for ADAS functions and present the results according to their hardware support to show experienced researchers the state-of-the-art institute results and provide them with the future development possibility. By making comparison of the automated driving classification from different areas, we raise some challenges and future work in the development of ADAS. And the deployment of ADAS in current commercial markets are also concluded in our paper to show the current ADAS application situations

\begin{figure*}[]
\centering
\includegraphics[width=\textwidth]{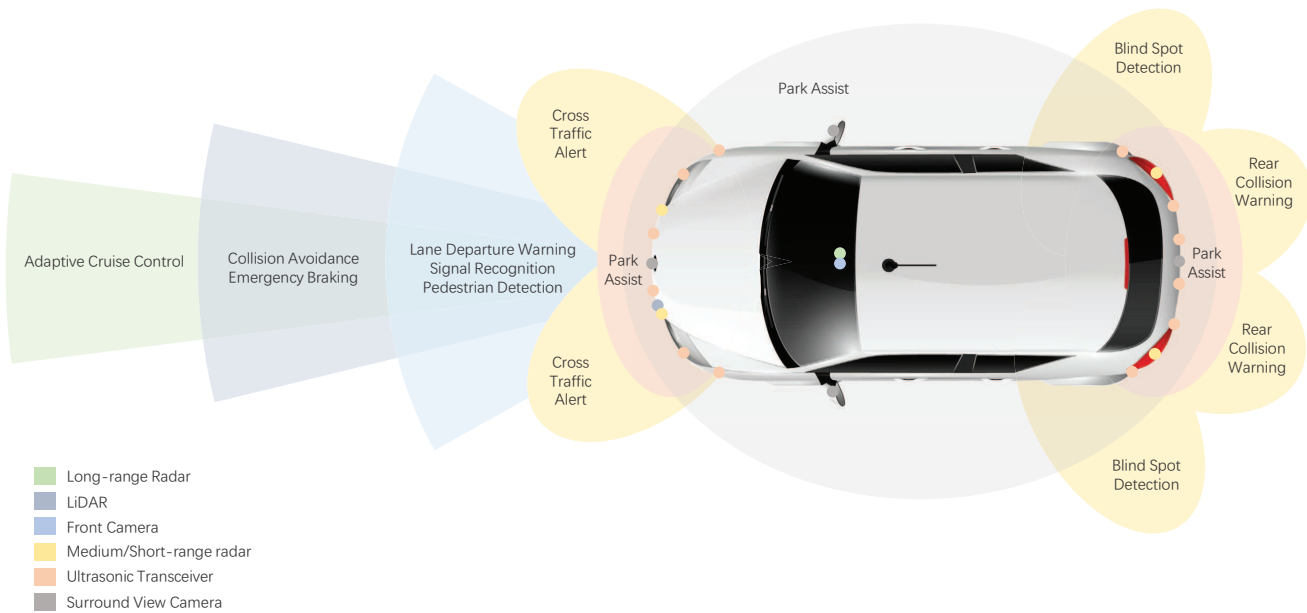}
\caption{Perception sensor location and corresponding ADAS function support.}
\label{f1}
\end{figure*}

\section{Perception Sensors}\label{hardware}

This paper aims to make a detailed explanation about ADAS from hardware support, state-of-the-art research algorithms and commercial applications. The main contributions of this paper are summarized as follows. Firstly, this paper is based on specific ADAS functions rather than high level automated driving tasks. We hope to present ADAS related knowledge clearly for newcomers and summarize current development focusing on ADAS only. Secondly, the introduction in this paper is specifically related to current commercial vehicles, instead of vehicles from institutions or laboratories, to better show current level of ADAS and actual extent. For example, all the sensor information is presented with their practical application in commercial vehicles such as installation locations, category selection, pros and cons for applied ADAS functions. Thirdly, most of collected papers were published in recent three years, so that the algorithms are all the most state-of-the-art and of great reference value for future development.

The structure of the paper is organized as follows. Section II makes an introduction about conventional perception sensor of ADAS, such as LiDAR (light detection and ranging sensor), camera, radar, ultrasonic transceiver and some other localization sensors. Their usual locations on the vehicle, related ADAS functions together with their advantages and disadvantaged are all included in this section. Section III reviews the latest researches about traffic sign detection, road detection system and moving object detection, applying the sensors mentioned in Section II. In Section IV, the deployment of ADAS in different car brands are compared to show the discrepancy of the application of ADAS in various areas.

Perception, as the first step of ADAS, refers to collect environmental information and provide useful surrounding messages for drivers to avoid possible risks and improve driving comfort \cite{pendleton2017perception}. It is achieved relying on different types of sensors mounted on vehicles. As can be seen, the common installation positions and corresponding ADAS function supports about exterior perception sensors such as LiDAR, camera, radar and ultrasonic transceiver are illustrated in Fig.1. Detailed implementation information and other related requirements about the above hardware are presented in the following content. Moreover, interior perception sensors for ADAS are also introduced in this section.

\subsection{LiDAR}\label{lidar}

LiDAR is an important segment of autonomous vehicle development. It is mainly used for real-time vicinity perception and high-resolution mapping. This section introduces LiDAR sensor from its related ADAS features including the usage for perception and localization. Its practical installation and selection methods inside the vehicle are also presented to make a brief summary for specific ADAS functions. Besides, the strengths and shortcomings are listed at the end of this part.

The first notable application of LiDAR was in the 2005 Grand DARPA Challenge \cite{thrun2006stanley}, where the winner (Stanley) mounted $5$ LiDAR on the roof to measure and detect forward as far as $25$ meters. Since the great performance of LiDAR in that race, it has caught great attention and become one of the most significant part of the advanced driving sensor suite.

In ADAS, the most important role of LiDAR sensor is for object detection, e.g., collision avoidance, pedestrian detection and park assistance. LiDAR for short range perception like forward collision prevention systems uses inexpensive low-range and low-resolution versions to measure the range between $1$ and $10$ meters \cite{bengler2014three}. It is also implemented in some low speed scenes, such as parking assistance and emergency brake assist \cite{ziebinski2016survey}, which are all the key features in the commercial vehicle development plans like ``City Safety'' (Volvo 2014) and ``City Stop'' (Ford 2014). For long range detection as pedestrian recognition and object tracking, LiDAR can provide accurate range information and larger field of view, since the emitted laser light can reach up to $200$ meters before reflection and the $3D$ point clouds it builts help for better segmentation \cite{wang2019pseudo}. The sensor can distinguish between obstacles for which a vehicle needs to stop such as pedestrians who are crossing a street and those where a vehicle does not need stop such as rising emissions, road damage.

Another common usage of LiDAR sensor in ADAS is localization. LiDAR-based localization methods \cite{badue2020self} offer measurement accuracy and easiness of processing. The reflectance intensity distribution of environment measured by LiDAR scanner can build the $3D$ object image in the environment, which can be further processed for object recognition or motion prediction.

Currently, LiDAR sensors are installed around the vehicle or above the vehicle. At present, most autonomous vehicle R$\&$D Labs have chosen the same installation way as Grand DARPA Challenge to equip LiDAR sensor on the roof of the car. This kind of sensor is mostly multi-line LiDAR (usually more than $16$-$lines$), which can scan the surrounding in $360^{\circ}$, to make long-distance detection and precise mapping for $3D$ road condition. Another way to assemble LiDAR sensor is installing them around the car. This kind of sensor, usually single-line LiDAR or LiDAR under $8$-$lines$, can only detect in partial angle, but it is much more convenient to install them inside the vehicle rather than expose directly outside the body shell. Audi released $A8$ Sedan equipped with Valeo's Scala LiDAR in late 2017, which is the first mechanical spinning LiDAR equipped into mass-produced commercial car.

LiDAR is a well-established tool in ADAS in poor light condition. LiDAR uses shorter wave length and superior beam properties than the traditional radar sensor, which offers a more suitable choice for $3D$ imaging and point cloud generation \cite{behroozpour2017lidar}. However, LiDAR may present a bad behavior in poor weather conditions such as heavy rain, snow, or fog, since these kinds of weather influence light reflection and refraction. Present LiDAR sensor has a steep price tag which also affects its commercial promotion in ADAS market.

Although the traditional electro-mechanical LiDAR mentioned above is able to rotate and scan the surrounding in a quite long range, they are bulky in design and expensive in sale. In this case, some other state-of-the-art methods like solid state LiDAR \cite{poulton2017coherent} come out and integrate the construction of LiDAR sensor on one chip. Since no moving mechanical parts are involved, it is more resilient in vibration, space-saving in construction and even has higher resolution and scanning rate.

\subsection{Camera}\label{cam}

This section makes a brief representation about camera. Different types of camera served for ADAS are introduced according to their installation positions. Some other special-used cameras are presented by analyzing their specific features. In the end, pros and cons about camera are concluded.

Camera is one of the leading sensors for ADAS due to its wide perception of the environment and acceptable price. The capability of acquiring the texture and color of all objects \cite{oniga2009processing} at the same time makes camera especially suitable for road surface estimation and traffic signal detection. If it is used in pairs, as stereo vision system, camera system is also capable of measuring distance, which can be further processed for localization and mapping.

According to the location of cameras in the vehicle, it can be summed as exterior camera and interior camera, which focus on surrounding environment perception and driver state detection respectively. The exterior cameras can be subdivided as the front camera and the surround view camera served for different ADAS functions.

Front cameras are usually attached behind the top of windshield and provide long-range visibility to detect forward situation and distant objects. Many basic ADAS functions are achieved based on front cameras. Forward collision warning \cite{galvani2019history}, as one of the most basic assistant applications, uses front cameras to detect obstacles, refine their relative position and classify them into cars, tracks or pedestrians. Traffic sign recognition informs drivers with current speed limits, road rules and warnings with the color and shape recognition capability of cameras. Lane departure warning (LDW) is realized with front cameras by capturing ground information in real-time and distinguishing the lane marker using computer vision algorithms. These features can be further applied in lane keeping system and lane centering assistant function with the help of mechanical constructions.

Surround view cameras, mounted around the vehicle, are usually short range cameras. To provide $360^{\circ}$ bird's eye view for the car, the surround view system sensor suite normally consists of four wide-angle cameras, mounted i.e. at the front bumper, rear bumper and the other two under each side mirrors \cite{zhang2014surround}. This kind of cameras works at low frame rate. In this case, it is designed for some low-speed situation like parking. After the image geometric and photometric alignment \cite{appia2015surround}, there will be a bird's eye view providing for drivers as park assistance function. Moreover, the surround view system also benefits in blind spot monitor function by detecting the surrounding objects approaching the vehicle.

The interior camera is usually installed at the top of the instrument cluster, detecting driver's head pose and tracing eye information to monitor the driver's state. It is designed to identify driver state condition when the driver's head changes abnormally or the driver's gaze is not facing forward to achieve driver drowsiness detection \cite{yoo2020optimization}. The advanced function of interior camera is to distinguish whether the driver is drunken or not, which makes pivotal contribution to road safety.

Besides the conventional monocular camera mentioned above, there are some other special cameras for specific functions. When cameras are used in pairs, as stereo camera, due to the perspective differences between the two camera images, the movement of the front objects and their distance can be measured within a range of $20$-$30$ meters \cite{ziebinski2016survey}. Therefore, in the vehicle manufacture, stereo camera can be used as a substitute of LiDAR sensor with lower price but also smaller field of view. Thermal camera is also widely used in ADAS. It can detect anything that generates or contains heat which shows a great performance in night vision or in bad weather conditions. Furthermore, in daytime driving, with the help of thermal cameras, the image capture can be better processed to reduce the redundancy from visible cameras.

Camera is widely used in ADAS functions for many reasons. It is small in size so that the freedom of installation is high. Due to the color capture capability of cameras, the specific object recognition like traffic lights and gesture can be distinguished with fewer post process. Moreover, with the support of deep learning, high-resolution image collected by cameras can better present the surrounding information. There are still some weak points of camera sensor, since the traditional monocular camera is sensitive to lighting and weather condition. And the extraction of this kind of information requires a complex and heavy computational process \cite{vivacqua2017low}.

\subsection{Radar}\label{radar}

Radar is introduced from three classic types in this section. The basic knowledge and inherent features about radar are firstly presented. And then their suitable ADAS functions are proposed. Benefits and weaknesses about radar are summed at last, together with the potential future challenges for radar applications and development.

Radar, as the standard configuration in ADAS suite, is the first sensor applied on the vehicles among all the perception hardware \cite{patole2017automotive}. It plays an important role in distance measurement and relative velocity detection by emitting electromagnetic waves and receiving reflections.

In recent years, radar is usually used coupled with camera or LiDAR to make up for the loss caused by blind spots or environmental conditions ensuring driving safety and ride comfort. It can be classified into short-range, medium-range and long-range installing at different locations for diverse ADAS functions.

Long-range radar, which is capable to reach objects up to $200$ meters, is always fitted at the middle of the front bumper. It can detect from $10$-$250$ meters at the wide-range of $\pm15^{\circ}$ and recognize multiple objects simultaneously, which is particularly suitable for long distance forward obstacles detection and collision avoidance. Since the ability of remote obstacle recognition prepares enough range and time for braking, it is used in ADAS like automatic emergency braking and traffic jam assistance \cite{fan2020computer}, which guarantees high speed safety. By measuring the reflection of electromagnetic waves, radar can also help in relative speed calculation, which ensures the adaptive cruise control (ACC) function.

Medium-range radar, with field of vision to about $60$ meters and $\pm40^{\circ}$, is fit for pedestrian detection and blind spot warning (BSW) \cite{galvani2019history}. It is symmetrically mounted below each headlight to monitor the environment behind or next to the vehicle. In this case, medium-range radar is used in lane departure assistance system by detecting the blind spot around the host vehicle and warning the driver in rear-facing mirrors. It can also be used in cross-traffic alert systems, when there are cyclists or pedestrians approaching the vehicle from blind spots.

The widest field of view among radar is short-range radar. It is capable to detect from $0.5$-$20$ meters with the wide of $\pm80^{\circ}$. Most short-range radars are placed at the corner of the front and rear bumpers and some vehicles even have short-range radars on both side of doors \cite{preussler2019photonically}. The main ADAS function of short-range radar is to support parking assistance system to alert the driver when reversing out from a parking space.

As the key sensor of ADAS system, radar shows a unique performance at a reasonable price \cite{feng2020deep}. It is robust to inclement environments and insensitive to lighting or weather change. With the Doppler effect, radar signal is easier to make a distinction between still and moving objects. Compared with LiDAR, it shows better detection capability, since the electromagnetic waves can see through many obstacles and feedback with more environment information. On the other side, the signal of automotive radar is usually in low resolution \cite{ziebinski2016survey}, which makes it challenging in object classification. Owing to the reflection process, it can be inaccurate in curved object representation and there might be loss of signal in multi-path reflection. With the rapid development of ADAS system and the popularity of intelligent vehicle, the issue about the same frequency radar signal interference has also caught extensive attention, which is also one of an urgent problem to be solved \cite{aydogdu2020radar}.

\subsection{Ultrasonic Transceiver}\label{ut}

Ultrasonic sensor is the basic and numerous perception sensor mounted on the vehicle, owing to the lower price and robust performance. It transmits over $2000Hz$ frequency magnitude signal and calculates the source-object distance when receiving the echo \cite{fan2020computer}.

This kind of sensor is mainly used in parking assistance system, since it is suitable for near-range detection, which reaches up to only $10$ meters and it prefers to work in low speed scenarios \cite{feng2020deep}, because of the sound signal attenuation. In 2016, Tesla released Model S with the new version of Autopilot, which provides automatic ``Reverse Park Assist'' function. This vehicle is equipped with $12$ ultrasonic sensors and each of them is placed on the corners of the car body to offer wide range perception \cite{ingle2016tesla}. Based on the parking zone surrounding information from ultrasonic sensors, the vehicle can successfully park in parallel or into the garage. Moreover, ultrasonic sensor can also be used in some ADAS functions like blind spot detection or collision avoidance to detection obstacles and warn the driver in real-time.

Ultrasonic sensor is generally the cheapest among all the sensor and independent from atmospheric conditions or light level, which results in the wide use in ADAS. But the slow responding time \cite{alonso2011ultrasonic} and only one-dimensional distance measurement \cite{rhee2019low} limit the further implementation of ultrasonic sensor. The low resolution leads to the inability of small object detection and precise target feature extraction. The noise signal interference and sound wave disturbance are also difficulties in this field for further analysis.

To make better choices for each ADAS function in sensor selection, features about the above four exterior sensors are concluded and compared in Fig.2. As can be seen, LiDAR is the most expensive perception sensor and requires for large computation cost, but it shows the best performance in all other features. Camera is suitable for perception in almost all fields, for this reason, it is the most widely used sensor in current ADAS market. Long-range radar is specific useful in distance detection, while ultrasonic sensor is the cheapest, which is affordable for vehicle of different qualities.

\subsection{Localization Sensors}\label{loc}

In addition to the traditional exteroceptive sensors mentioned above, there are some other types of sensors focusing on geolocation information collection, which helps for environment detection, vehicle motion decision and future path planning. Global positioning system (GPS) and inertial measurement unit (IMU) are representatives of this type of sensors.

GPS works by receiving data from more than three satellites to figure out the current time and position \cite{luo2019localization}. This kind of information can be utilized for vehicle navigation. The advantage of using GPS is that with the existing maps, the approximate location of a vehicle can be located. However, GPS is only accurate to around $1$-$2$ meters, and it requires map data updating in time \cite{kastrinaki2003survey}. Moreover, the signal from satellites can be very weak in indoor areas such as tunnels, and high-rise buildings in a city may also influence signal receiving \cite{liu2017computer}. IMU sensor provides internal movement information by measuring a vehicle's acceleration along $X, Y, Z$ axes. It shows the vehicle motion trace rather than self-location information and the accuracy degrades along the time, due to error accumulation \cite{campbell2018sensor}. However, integrating those localization sensors with novel mapping sensors like LiDAR can help to reduce positioning cost to a certain extent \cite{zekavat2011handbook}, improve the accuracy to only a few centimeters, and relive from the computation pressure of real-time calculation.

\begin{figure}
\centering
\includegraphics[width=2.5in]{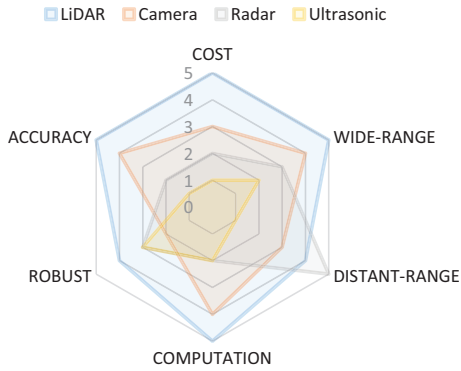}
\caption{The comparison of sensor features.}
\label{f2}
\end{figure}

\begin{table*}[]
\centering
\caption{ADAS Application Summary}
\label{support}
\resizebox{\textwidth}{!}{%
\scriptsize
\begin{tabular}{@{}llcccc@{}}
\toprule
\textbf{Classification in this paper}     & \textbf{ADAS functions}  & \textbf{Camera} & \textbf{LiDAR} & \textbf{Radar} & \textbf{Ultrasonic} \\ \midrule
Traffic Sign Detection                    & Traffic Sign Recognition & \checkmark               &                &                &                     \\
\multirow{5}{*}{Road Detection System}    & Lane Departure Warning   & \checkmark               &                &                &                     \\
                                          & Lane Centering           & \checkmark               &                &                &                     \\
                                          & Lane Keeping Assistance  & \checkmark               &                &                &                     \\
                                          & Adaptive Cruise Control  & \checkmark               &                & \checkmark              &                     \\
                                          & Intersection Assistance  & \checkmark               &                &                &                     \\
\multirow{7}{*}{Moving Object Detection}  & Collision Avoidance      & \checkmark               & \checkmark              & \checkmark              &                     \\
                                          & Parking Assistance       & \checkmark               &                & \checkmark              & \checkmark                   \\
                                          & Pedestrian Detection     & \checkmark               & \checkmark              &                &                     \\
                                          & Cross Traffic Alert      & \checkmark               &                & \checkmark              & \checkmark                   \\
                                          & Emergency Braking System & \checkmark               & \checkmark              & \checkmark              &                     \\
                                          & Blind Spot Detection     &                 &                & \checkmark              & \checkmark                   \\
                                          & Rear Collision Warning   &                 &                & \checkmark              &                     \\ \bottomrule
\checkmark: This sensor is usually applied for the forward ADAS function.
\end{tabular}%
}
\end{table*}

\section{Prediction Algorithms}\label{Algorithms}

In this section, state-of-the-art researches about ADAS implementation methods are presented. These algorithms use raw data collected directly from the above perception sensors and obtain relevant features for ADAS to achieve the purpose of assisting or warning drivers.

ADAS functions can be classified into several sections, according to their oriented approaches and practical driving assistant efficacies \cite{marti2019review}. In this paper, ADAS functions are concluded into three main types as in Table II, based on detailed ADAS targets. Supporting hardware sensors for each function are also marked in the above table. In this case, the following subsections are organized to explain the implementation algorithms about ADAS based on perception sensors in detail.

\subsection{Traffic Sign Detection}\label{TSD}

Traffic sign detection, which has raised much attention in recent years, is an inevitable task for ADAS and intelligent vehicles. Since traffic signs present with different colors and shapes, there are many solutions to make detection and classification. In this section, recent works about traffic sign detection are categorized based on the sensor usage.

\subsubsection{Camera-Based}\label{TSDcam}

Due to the advanced computer vision technology and the efficient image processing, camera is already used as the main sensor for traffic sign detection. The pattern recognition is realized based on traditional methods like color, Histogram of Oriented Gradient (HOG) or edge feature extraction or machine learning \cite{mogelmose2012Vision}.

\cite{he2019traffic} has raised a semi-supervised learning approach and used limited numbers of labeled signs to train a large number of traffic signal recognition. They extract signal features from color, edge features, HOG features, and train models based on the above three tags using different feature sets. Samples with high confidence are added into labeled data set for following convolutional neural networks (CNN) model training and final classification results come from the Adaboost algorithm.

The machine learning-based method is regarded as a fast and simple recognition solution. In this case, it is attractive to find out the most suitable framework via methods comparison. In \cite{arcos2018evaluation}, authors have made an algorithm comparison of convolutional neutral networks like region based CNN (R-CNN), region based fully convolutional network (R-FCN), single shot multibox detector (SSD) and you only look once (YOLO), based on their performance in object detection using color image. They present their final result referring to the metrics, namely mean average precision, memory usage, running time, number of parameters, and so forth, and give recommended algorithms based on different requirements. \cite{arcos2018deep} has compared several stochastic gradient descent optimization algorithms and analyzed their influence on CNN about the traffic signal detection work.

Owing to the depth information limitation of cameras, traffic sign can be very small in images. In this case, many researches have been made specifically on small pattern detection in traffic recognition task. \cite{liang2019traffic} has tried to tackle small size sign detection using ResNet-50 network to build a pyramidal framework and combined features extracted from different layers to obtain high-level semantic feature maps in all scales. In \cite{zhang2020cascaded}, a multi-scale cascaded R-CNN has been presented to tackle the traffic signal recognition problem in small size and low resolution under bad weather condition. \cite{liu2020small} proposed a method using diverse region based CNN (DR-CNN) concatenating features from both shallow layers and deep layers to improve the accuracy of small traffic sign detection. The fused feature map of different layers can provide higher resolution and larger semantic information, and the point-wise convolution in this algorithm help weaken invalid background noise, which all benefit in small pattern detection.

\subsubsection{Sensor-Fusion}\label{TSDsf}

The detection only based on cameras is limited by many real-life factors like weather condition, shadows of surrounding objects, environment light and so on. The fusion of other sensors is for this reason necessary to improve robustness and reliability of traffic sign detection task.

Owing to the limitation of monocular camera in position collection, \cite{guan2018robust} has built a novel system integrating both mobile LiDAR and digital images. Mobile LiDAR is mainly used for signal position detection by figuring out their point cloud features according to prior knowledge of traffic poles. After the confirmation of signal position, segmentation of signal point cloud is projected onto digital image for content classification. A convolutional capsule has been used in \cite{guan2019convolutional} to recognize different types of traffic signs.

\cite{balado2020novel} has fused LiDAR sensor, $\pm360^{\circ}$ camera with GPS based on mobile mapping system to locate traffic signs. The panoramic video captured by camera is processed for traffic signal detection as the first step, since image processing technique is much quicker than other related methods. Other sensors are used for traffic position detection and collation to avoid false positives.

Apart from traffic signs recognition, the detection of traffic light state is also an important task for ADAS. Unlike variable types and different locations of traffic signs, traffic lights are usually built on the cross and generally include three visible states. In this case, \cite{hirabayashi2019traffic} has presented a traffic light recognition method based on camera detection and $3D$ map information. With the help of other self-location hardware embedded on the host vehicle, the location of surrounding traffic lights can be obtained previously. The region of interest about traffic lights in the image can then be fed into CNN for further classification.

\subsubsection{Other Related Works}\label{TSDwork}

Since all the traffic signs are placed outdoor, there is definitely some degradation on the signs like losing colors or out of shape owing to poor maintenance. These kinds of problems may influence the accuracy of traffic sign detection. For this reason, \cite{mannan2019classification} has proposed a novel methodology to solve degraded traffic sign recognition. They use a novel flexible linear mapping technique to first fix outer rim color fade problem and then send the output into a flexible Gaussian mixture dynamically updating split and merge scheme to capture the feature of the original signal. It shows a superior outcome compared with traditional hand crafted feature method, which can greatly improve the performance of traffic sign detection technology in actual application scenarios.

The images captured directly from camera also include a lot of unrelated pixels related to the background environment, which will cause large computation cost and even a reduction of accuracy. To help build a slimmer and more accurate traffic sign classification system, \cite{zhang2020lightweight} has presented a two-lightweight convolutional neural network, which is specially suitable for intelligent mobile vehicle system. The first one focuses on feature distillation, which serves for the second traffic classification network. In this case, there will be fewer parameters and lower computations in the second network than that of the first one.

\subsection{Road Detection System}\label{RDS}

Many ADAS functions can be included into road detection system like lane departure warning, lane centering, lane change assistance, adaptive cruise control, intersection assistance, and so on. Among them, the perception of road scenarios and detection of painted lane markers are the principal tasks and they are vital challenges for safety insurance. In this section, road detection tasks are introduced according to camera-based and sensor fusion approaches.

\subsubsection{Camera-Based}\label{RDScam}

Most of current mature road detection systems are based on vision technology, owing to extensive computer vision knowledge background and cost-effective character of cameras \cite{xing2018advances}.

To improve road detection methods based on camera and extend classification image dataset in this field, \cite{chen2019deep} has raised a large-scale dataset named DrivingScene containing traffic scenarios under different weather condition, road structure driving locations and so forth. Based on this dataset, they have also proposed a novel multi-label neural network for category prediction learning and supervised learning. Their network can extract features from different sizes of inputs and reserve the most of image information.

The deep learning algorithm is the most popular choice for image process and the application of deep learning in traffic detection shows a rapid increase compared with conventional approaches. Among them, the convolutional neutral network is most commonly used thanks to its powerful extraction ability and superior robust performance. \cite{bayoudh2020transfer} has presented a hybrid $2$D-$3$D CNN model based on transfer learning to make the semantic detection of road space. \cite{guindel2019traffic} has rebuilt the surrounding $3D$ construction with the high-density information from stereo camera for scene detection by putting point clouds feature into a pre-trained mode to make scene classification.

For painted lane detection researches, \cite{ye2018lane}  has proposed a novel lane detection method by transferring the image lane structure into a waveform, which strengthens the lane marking features and helps to simplify the detection process when approaching or departing lane markers. When it comes to complex road conditions, the machine learning method is used with pretreated collected image data to eliminate non-lane regions before the generation of waveforms.

Due to the movement of vehicles and the possible influential environment factors, \cite{zou2019robust} has proposed a multiple of continuous driving scenes detection to reduce false negatives of lane detection via a hybrid deep neural network. It has combined deep convolutional neural networks (DCNN) and deep recurrent neural networks (DRNN), in which DCNN is used for semantic segmentation manner and DRNN shows its time-series analyzing capability to process with continuous image frames. This system specially suits for lane detection in real-world since the time-series processing offers much more information for lane feature extraction and host vehicle motion prediction.

Considering structural features in visual signal detection, \cite{li2016deep} has introduced a multi-task deep convolutional network by integrating CNN and RNN detectors. CNN is responsible for lane structure modeling using pre-trained geometric information model, while RNN is applied for lane boundary detection. Moreover, due to the prediction ability of RNN, this system can also be used to fix the hidden lane boundary caused by obstacles on the lane, which is useful in practical traffic scenes.

\cite{hou2019learning} has presented a self-attention distillation (SAD) method in a lane detection task. This model can learn from itself without any other addition label requirement and external supervision. It can be attached to any feed-forward convolutional neural networks to pre-train model preparing for lane detection without additional real-time computation cost. Similar to \cite{hou2019learning}, \cite{xiao2020attention} has presented an attention module by combining self-attention and channel attention. They utilize global and channel content together to distinguish whether a pixel belongs to lane marker or not, and in this case, realize the lane detection task.

\cite{song2018lane} has presented a lane detection method with stereo camera. The stereo image information is mainly used for obstacle segmentation and results are fed into self-adaptive traffic lane model for parallel lane detection. The self-adaptive traffic lane model is designed in Hough Space with maximum likelihood angle region of interest (ROI) and dynamic poles detection ROI. Based on the lane feature and range from host vehicle, detected lanes are classified to provide traffic rules for drivers.

\subsubsection{Sensor-Fusion}\label{RDSsf}

Finding out the region of roads can be much easier with the $3D$ environment structure information from LiDAR sensor.

In \cite{xiao2018hybrid}, a novel road detection method is introduced by fusing monocular camera and LiDAR sensor. A framework named conditional random field (CRF) is used to change road detection task into a binary labeling problem. The point cloud information and image pixels from the two sensors are labeled serving for road boundary detection. \cite{caltagirone2019lidar} has presented a novel fusion fully convolutional neural network and integrate camera image with LiDAR points clouds for road detection. The innovation in this system is training to make cross connections between the two sensor information processing branches in all layers, which shows the best performance among all other similar FCN systems.

Since the depth available and more robust information from LiDAR can help monocular camera to better detect and segment the environment content, many researches begin to integrate these two sensors together on road detection tasks. However, the simple sensor data fusion approaches, like directly projecting $3D$ LiDAR data into $2D$ image plane, can cause reduction of depth information, which makes road areas even less distinguishable in following process. \cite{chen2019progressive} has introduced a novel progressive LiDAR adaptation-aided road detection (PLARD) method to help LiDAR data more compatible with image information. This approach uses altitude difference-based transformation to align $3D$ data with perspective view and applies cascaded fusion structure for feature adaptation. Through the two adaptation modules, they exploit most of information from these sensors and largely improve the robustness of road detection in urban scenes.

\subsection{Moving Object Detection}\label{MOD}

Moving object detection is one of the most significant tasks for ADAS. It is responsible for surrounding environment detection and object classification. Object position, moving speed and even motion intention are also sometimes required to be figured out in this subsystem. These information analyzed from sensors ensures driving safety and works for ADAS like collision avoidance, parking assistance, pedestrian protection, cross traffic alert and so on. According to hardware application, it can be categorized into camera-based, LiDAR-based and sensor fusion methods.

\subsubsection{Camera-Based}\label{MODcam}

With direct color view capture ability, camera is always regarded as the most traditional and suitable sensor for object detection. Although there are some limitations about usage condition of camera, some new methods are proposed to improve camera behavior in this task.

To overcome low-resolution capture feature of visible-light camera at a dusky lighting condition, \cite{kim2018pedestrian} has used faster R-CNN in their present. The modified faster R-CNN is trained and a random additive white gaussian noise is added in this system for better robust performance against noise and illumination. In the practice test, the weighted summation of successive frame features is added in this application to provide temporal information.

\cite{li2019deep} has come up with three novel deep learning solutions based on YOLO to make up for limitation of camera performance in bad weather condition, like hazy days. A weighted combination layer is introduced in their presentation to raise pedestrian detection accurate output. To relieve computation pressure, linear bottleneck and depth wise separable convolution are used, which can lead to lower computation cost and fewer parameters.

Due to distinctiveness of human-like object detection, thermal sensor shows a superior performance in long-distance and low visibility environments. \cite{guan2019fusion} has introduced a novel pedestrian detection framework, which shows great performance in all day lighting condition using camera information only (monocular and thermal camera). The multi-spectral information under daytime and nighttime is firstly collected and is put into two-stream deep convolutional neural networks to extract data features under different illumination conditions, which benefits in the semantic segmentation and pedestrian detection.

In fact, the velocity of object is also possible to be obtained once the detection and classification process are finished. The base of velocity detection is the acquisition of depth information. A dynamic simultaneous localization and mapping (SLAM) algorithm is presented in \cite{henein2020dynamic}, which extracts velocity information using a RGB-deep (RGB-D) camera. Semantic segmentation in this experiment is used for rigid object motion estimation without the requirement of any other prior knowledge, which greatly helps the detection of object speed and provides assistance in implementation tasks, like collision avoidance and adaptive cruise control.

\subsubsection{LiDAR-Based}\label{MODli}

Traditional camera is unable to obtain distance information and it is also challenging to detect small objects on image owing to the limited resolution. In this case, LiDAR is another suitable alternate for camera in object detection task. The larger field of view and higher density environment information make LiDAR a right detection sensor.

\cite{wang2017pedestrian} has showed a traditional method for pedestrian recognition using $3D$ LiDAR. The $3D$ point cloud data collected from LiDAR is firstly projected on the $2D$ plane and then is computed by a support vector machine (SVM) pre-trained classifier for pedestrian selection. Once it is recognized as a pedestrian, the information about initial $3D$ points cloud position and velocity of host vehicle will be considered together for pedestrian tracking. The surrounding information like curb position and road width will be taken into account to avoid the possible collision accident.

\cite{ali2018yolo3d} has also proposed a real time object detection and classification system using LiDAR sensor alone. It is an end-to-end training approach based on extending YOLO-v2. The $3D$ LiDAR point cloud data is first projected into two bird's eye view grid map for height and density feature extraction and then is fed into this framework for $3D$ bounding boxes marking for object detection. With information from LiDAR point cloud, the system present acceptable accuracy and real time performance.

\cite{zhou2018voxelnet} has made a great contribution and breakthrough in the LiDAR usage. A novel architecture for object detection based on $3D$ LiDAR data is introduced in this paper. It unifies feature extraction from point cloud data and $3D$ bounding boxes, and make together in a trainable deep network named VoxelNet, which simplifies the procedure of $3D$ detection since there is no need to extra project $3D$ points cloud into $2D$ plane. It makes the most of sparse point structure feature of LiDAR sensor and the parallel processing method of VoxelNet raise efficiency of computation. Based on the superior $3D$ object detection result, it shows encouraging results in classification of objects, such as pedestrians and cyclists.

\cite{yang2020multi} has designed a multi-view semantic learning network (MVSLN) based on LiDAR data. The multiple view generator (MVG) module projects $3D$ point cloud into bird's eye view, rotated front view, rotated left view and rotated right view, which ensures the preservation of low-level feature. Spatial recalibration fusion module is used for four views' alignment to prepare for object detection in the following $3D$ region proposal network (RPN) module.

\subsubsection{Sensor-Fusion}\label{MODsf}

In order to integrate the advantages of different sensors and present more accurate real-time results, sensor fusion methods are put forward.

\cite{asvadi2018multimodal} has presented a real-time multi-modal vehicle detection methodology together with color camera and $3D$ LiDAR. LiDAR, as the sole sensor in this present, is applied for providing dense-depth map and reflectance map. Camera, in this design, is used for the alignment of $3D$ LiDAR to reduce the misdetection rate. The three individual data modalities serve as the input of a Deep ConvNet-based framework to generate bounding boxes of each vehicle in the detection results. This decision-level fusion surpasses a lot compared with individual modality detectors, which shows superior performance on road vehicle perception.

Sensor fusion detection method of \cite{gao2018object} is based on CNN and image upsampling theory. By extracting depth features from point cloud of LiDAR data, the greatest limitation of RGB camera can be remedied.. RGB data together with depth information is sent into a deep CNN for feature learning using purely supervised learning to make object detection and classification. Compared with traditional methods that use camera or LiDAR only, the sensor fusion approach shows higher accuracy and superior classification performance.

To make better use of LiDAR sensor, \cite{zhao2020fusion} has raised a complementary object identification approach with the fusion of $3D$ LiDAR and vision camera. To reduce processing time, object-region from $3D$ spatial information is firstly extracted using $3D$ LiDAR in three steps: figuring out ground area from $3D$ point cloud and making the removal, segmenting the rest non-ground point into isolation and then generating the final region proposal, which can be used as the inputs of CNN model. The object classification result is presented after the feature extraction in CNN.

\cite{chen20173d} has realized object detection with the fusion of $3D$ LiDAR and stereo camera rather than the traditional vision camera, since the depth and stereo information from camera can improve $2D$ bounding boxes to $3D$, which helps to compute efficiently. The $3D$ features can be further extracted by structured SVM and the outputs are fed into a $3D$ object detection neural network to predict $3D$ bounding boxes of objects. These high-quality $3D$ object detection results show better classification performance in CNN model.

\section{The Application of ADAS}\label{use}

Although hardware performance like accuracy and implementation capability for ADAS keeps improving, sensor installation on vehicles is all in readiness, and the related technologies develop quickly, it is still challenging for further promotion on ADAS deployment. Therefore, it is necessary to make corresponding localization adjustments for the regions where vehicles are actually used in order to fit for specific weather conditions, traffic features and humanistic environments.

\begin{table*}[]
\centering
\caption{ADAS Support in Commercial Vehicles}
\label{brand}
\resizebox{\textwidth}{!}{%
\begin{tabular}{ccccllccl}
\hline
\multirow{2}{*}{\textbf{Brand}} & \multirow{2}{*}{\textbf{Country}} & \multirow{2}{*}{\textbf{Model}} & \multicolumn{5}{c}{\textbf{Sensors}}                                                                                                                                                                                                                                                           & \multicolumn{1}{c}{\multirow{2}{*}{\textbf{ADAS Features}}}                                                                                                                                                     \\ \cline{4-8}
                                &                                   &                                 & LiDAR                     & \multicolumn{1}{c}{Camera}                                                                          & \multicolumn{1}{c}{Radar}                                                                              & Ultrasonic & GPS                                    & \multicolumn{1}{c}{}                                                                                                                                                                                            \\ \hline
Tesla                           & USA                               & Model S (AUTOPilot)             & -                         & \begin{tabular}[c]{@{}l@{}}3 Front \\ 2 Side \\ 2 Rear\\ 1 Backward\end{tabular}                    & 1 Forward-facing Long-range                                                                            & 12         & \checkmark              & \begin{tabular}[c]{@{}l@{}}Emergency Braking System\\ Parking Assistance\\ Lane Change Assistance\\ Lane Departure Warning\end{tabular}                                                                         \\
Ford                            & USA                               & Escape (Co-Pilot360)            & -                         & \begin{tabular}[c]{@{}l@{}}1 Front\\ 1 Reverse Camera\end{tabular}                                  & \begin{tabular}[c]{@{}l@{}}1 Forward-facing Long-range\\ 2 Short-range\end{tabular}                    & 12         & \checkmark              & \begin{tabular}[c]{@{}l@{}}Blind Spot Detection\\ Adaptive Cruise Control\\ Lane Centering\\ Lane Keeping\\ Post Collision Braking\\ Pre-Collision Assist Braking\end{tabular}                                  \\
Cadillac                        & USA                               & CT6 (SuperCruise)               & -                         & \begin{tabular}[c]{@{}l@{}}1 Forward\\ 4 Surround-View\\ 1 Driver (monocular\&thermal)\end{tabular} & \begin{tabular}[c]{@{}l@{}}1 Long-range\\ 5 Short-range\end{tabular}                                   & 12         & \checkmark(with HD map) & \begin{tabular}[c]{@{}l@{}}Lane Detection\\ Traffic Signal Recognition\\ Moving Object Detection\end{tabular}                                                                                                   \\
Benz                            & Germany                           & S Class                         & -                         & \begin{tabular}[c]{@{}l@{}}1 Front (stereo)\\ 4 Surround-View\end{tabular}                          & \begin{tabular}[c]{@{}l@{}}2 Front Multi-mode\\ 1 Front Long-range\\ 2 Rear Multi-purpose\end{tabular} & 12         & \checkmark              & \begin{tabular}[c]{@{}l@{}}Active Distance Assist\\ Active Lane Change\\ Active Speed Limit Assist\\ Emergency Stop Assist\\ Blind Spot\\ Lane Keeping\\ Traffic Sign Recognition\\ Active Parking\end{tabular} \\
BWM                             & Germany                           & 7 Series(Pro)                   & -                         & \begin{tabular}[c]{@{}l@{}}3 Front\\ 4 Surround-View\\ 1 Driver (thermal)\end{tabular}              & \begin{tabular}[c]{@{}l@{}}1 Long-range\\ 4 Short-range\end{tabular}                                   & 12         & \checkmark              & \begin{tabular}[c]{@{}l@{}}Active Cruise Control\\ Collision Warning\\ Lane Departure Warning\\ Dynamic Brake Control\\ Lane Keeping Assistant\\ Lateral Collision Avoidance\end{tabular}                       \\
Audi                            & Germany                           & A8                              & \checkmark & \begin{tabular}[c]{@{}l@{}}1 Front\\ 4 Surround-View\\ 1 Thermal Front\end{tabular}                 & \begin{tabular}[c]{@{}l@{}}1 Long-range\\ 4 Medium-range\end{tabular}                                  & 12         & \checkmark              & \begin{tabular}[c]{@{}l@{}}Adaptive Cruise Control\\ Turn Assist\\ Adaptive Cruise Assist\\ Active Lane Assist\\ Collision Avoidance Assist\\ Traffic Sign Recognition\end{tabular}                             \\
Volvo                           & Sweden                            & XC90                            & -                         & \begin{tabular}[c]{@{}l@{}}1 Front\\ 4 Surround-View\end{tabular}                                   & \begin{tabular}[c]{@{}l@{}}1 Long-range\\ 1 Medium-range\end{tabular}                                  & 12         & \checkmark              & \begin{tabular}[c]{@{}l@{}}Speed Limit\\ Distance Alter\\ Adaptive Cruise Control\\ Pilot Assist\\ Lane Keeping Aid\\ Passing assistance\\ Rear Collision Warning\\ Blind Spot Information\end{tabular}         \\
NIO                             & China                             & ES8 (NIOpilot)                  & -                         & \begin{tabular}[c]{@{}l@{}}3 Front\\ 4 Surround View\\ 1 Driver Monitoring\end{tabular}             & \begin{tabular}[c]{@{}l@{}}1 Medium-range\\ 4 Short-range\end{tabular}                                 & 12         & \checkmark              & \begin{tabular}[c]{@{}l@{}}Traffic Jam Pilot\\ Lane Keeping Assist\\ Cross Traffic Alert-Front\\ Traffic Sign Recognition\\ Advanced Parking\end{tabular}                                                       \\ \bottomrule
\end{tabular}%
}
\end{table*}

The current most popular classification about automated driving and ADAS is the taxonomy from SAE 2016, which determines driving automation level classification by distinguishing the charging role during the whole dynamic driving task between driving user and driving system. Moreover, in this recommended practice document, it has also raised some examples about specific ADAS features based on each level like ACC, LDW, BSW, and so on. This clear and structured taxonomy is definitely helpful for automobile manufacturers and it is also useful in the planning of the overall intelligent vehicle development. Actually, before the SAE taxonomy released in 2016, there are also some other simple reports or policies which are made to set simply definitions about levels of vehicle automation. In 2013, the national highway safety administration (NHTSA) of USA defined the levels of vehicle automation for the first time in the world for intelligent vehicle institutes, which made a preparation for automated vehicle development. Based on this, other hardware or electric elements standards were made to keep pace with the intelligent vehicle development. Related novel features added on vehicles were also equipped in their new car assessment program (NCAP) for safety test. However, descriptions of each level in this document are in loosely descriptive terms and the definition of each term is not quite specific. In late 2013, German federal highway research institute posted their standard in `Legal consequences of an increase in vehicle automation', which is performed based on German law, and it is also the main reference of SAE standard.

On basis of these classification standards, at the end of 2020, China published the national standard for the automatic classification of automated driving. Compared with previous classification standards, the taxonomy in this standard is product-oriented and mainly defines the classification of driving automation systems and model technical requirements for the system, rather than requirements for driving automation system users. Except for automatic level definition, the documents about detailed definitions of ADAS features and specific feature testing methods were all released, such as sections of intelligent vehicle innovation and development strategy in China. It is planned in the strategy that by 2025, technological innovation, industrial ecology, infrastructure, regulations and standards, product supervision and network security system of intelligent vehicles in China will be basically formed. Smart cars that is able to realize conditional autonomous driving can be produced in mass production, and smart cars that is able to realize highly autonomous driving can be marketed in specific environments. The concept of ``cloud" is proposed for the first time in this strategy, and it strives to realize a collaborative system between vehicles and other traffic participants, vehicles and vehicles, vehicles and road infrastructure, and vehicles and cloud service platforms. It not only formulates autonomous driving classification standards in China, which provides a basis for the subsequent promulgation of laws and regulations related to autonomous driving, but also provides guidelines for enterprises to develop autonomous driving, and helps accelerate the development of autonomous driving industries.

To make improvement of the ADAS deployment, in addition to specify automated standards according to each national condition, it is also important for vehicle companies to adjust the embedded ADAS features based on each region conditions. For example, cameras mounted outside vehicles may meet severe sunstrike due to sunlight angled almost horizontally in northern Europe especially during winter. And in Germany, owing to the lack of speed limit on parts of the autobahn, the speed differences between vehicles may become larger. Therefore, it is much challenging for features like speed detection and collision avoidance. Table III shows the specific ADAS features and the embedded hardware sensors of some famous vehicle companies from different areas. However, as can be seen, there are little differences in the hardware applications and providing ADAS features.

\section{Conclusion}\label{s7}

In this paper, basic perception sensors are introduced from their inherent features and the practical applications in commercial vehicles together with the supporting ADAS functions.  Current novel algorithms and researches for ADAS implementation are briefly presented according to traffic signal recognition, road field detection and moving object detection. In the last section of this paper, different versions of taxonomy about automated driving are listed and discussed. Corresponding projects and plans in China about automated driving and ADAS are presented in detail. This paper mainly focuses on ADAS features and introduces basic ADAS functions from both hardware supports and implemented algorithms, which provides a clear representation for newcomers in the automated driving field. We pay much attention on commercial vehicle construction and possible implemented researches, and also analyze the current ADAS application situation at the end of this paper. Through the efforts above, this paper is dedicated to show readers the mass production situation of ADAS and raise potential development for them.

For further development and wider deployment of ADAS, except for researches about hardware and algorithms, it is necessary to make standards and requirements combining local environment, and manufacturers shall adjust the basic ADAS features for region conditions.

%



\begin{thebibliography}{10}

\bibitem{zhu2017overview}
H.~Zhu, K.-V. Yuen, L.~Mihaylova, and H.~Leung, ``Overview of environment
  perception for intelligent vehicles,'' \emph{IEEE Transactions on Intelligent
  Transportation Systems}, vol.~18, no.~10, pp. 2584--2601, 2017.

\bibitem{TaxonomyAD}
SAE, ``Taxonomy and definitions for terms related to driving automation systems
  for on-road motor vehicles,'' SAEJ3016, Tech. Rep., 2016.

\bibitem{kuutti2020survey}
S.~Kuutti, R.~Bowden, Y.~Jin, P.~Barber, and S.~Fallah, ``A survey of deep
  learning applications to autonomous vehicle control,'' \emph{IEEE
  Transactions on Intelligent Transportation Systems}, vol.~22, no.~2, pp.
  712--733, 2021.

\bibitem{eskandarian2019research}
A.~Eskandarian, C.~Wu, and C.~Sun, ``Research advances and challenges of
  autonomous and connected ground vehicles,'' \emph{IEEE Transactions on
  Intelligent Transportation Systems}, vol.~22, no.~2, pp. 683--711, 2021.

\bibitem{national2020most}
NTSB, ``Most wanted list of transportation safety improvements,'' National
  Transportation Safety Board, Tech. Rep., 2020.

\bibitem{gordon2015automated}
T.~Gordon and M.~Lidberg, ``Automated driving and autonomous functions on road
  vehicles,'' \emph{Vehicle System Dynamics}, vol.~53, no.~7, pp. 958--994,
  2015.

\bibitem{buehler2009darpa}
M.~Buehler, K.~Iagnemma, and S.~Singh, \emph{The {DARPA Urban Challenge}:
  Autonomous Vehicles in City Traffic}.\hskip 1em plus 0.5em minus 0.4em\relax
  Springer, 2009, vol.~56.

\bibitem{broggi2010development}
A.~Broggi, P.~Medici, E.~Cardarelli, P.~Cerri, A.~Giacomazzo, and N.~Finardi,
  ``Development of the control system for the vislab intercontinental
  autonomous challenge,'' in \emph{13th International IEEE Conference on
  Intelligent Transportation Systems}, 2010, pp. 635--640.

\bibitem{geiger2012team}
A.~Geiger, M.~Lauer, F.~Moosmann, B.~Ranft, H.~Rapp, C.~Stiller, and
  J.~Ziegler, ``Team {AnnieWAY's} entry to the {2011 Grand Cooperative Driving
  Challenge},'' \emph{IEEE Transactions on Intelligent Transportation Systems},
  vol.~13, no.~3, pp. 1008--1017, 2012.

\bibitem{galvani2019history}
M.~Galvani, ``History and future of driver assistance,'' \emph{IEEE
  Instrumentation \& Measurement Magazine}, vol.~22, no.~1, pp. 11--16, 2019.

\bibitem{ziebinski2016survey}
A.~Ziebinski, R.~Cupek, H.~Erdogan, and S.~Waechter, ``A survey of {ADAS}
  technologies for the future perspective of sensor fusion,'' in
  \emph{International Conference on Computational Collective
  Intelligence}.\hskip 1em plus 0.5em minus 0.4em\relax Springer, 2016, pp.
  135--146.

\bibitem{lu2005technical}
M.~Lu, K.~Wevers, and R.~Van Der~Heijden, ``Technical feasibility of advanced
  driver assistance systems {(ADAS)} for road traffic safety,''
  \emph{Transportation Planning and Technology}, vol.~28, no.~3, pp. 167--187,
  2005.

\bibitem{yurtsever2020survey}
E.~Yurtsever, J.~Lambert, A.~Carballo, and K.~Takeda, ``A survey of autonomous
  driving: Common practices and emerging technologies,'' \emph{IEEE Access},
  vol.~8, pp. 58\,443--58\,469, 2020.

\bibitem{badue2020self}
C.~Badue, R.~Guidolini, R.~V. Carneiro, P.~Azevedo, V.~B. Cardoso, A.~Forechi,
  L.~Jesus, R.~Berriel, T.~M. Paixao, F.~Mutz \emph{et~al.}, ``Self-driving
  cars: A survey,'' \emph{Expert Systems with Applications}, p. 113816, 2020.

\bibitem{marti2019review}
E.~Marti, M.~A. de~Miguel, F.~Garcia, and J.~Perez, ``A review of sensor
  technologies for perception in automated driving,'' \emph{IEEE Intelligent
  Transportation Systems Magazine}, vol.~11, no.~4, pp. 94--108, 2019.

\bibitem{van2018autonomous}
J.~Van~Brummelen, M.~O¡¯Brien, D.~Gruyer, and H.~Najjaran, ``Autonomous vehicle
  perception: The technology of today and tomorrow,'' \emph{Transportation
  Research Part C: Emerging Technologies}, vol.~89, pp. 384--406, 2018.

\bibitem{ziebinski2017review}
A.~Ziebinski, R.~Cupek, D.~Grzechca, and L.~Chruszczyk, ``Review of advanced
  driver assistance systems {(ADAS)},'' in \emph{AIP Conference Proceedings},
  vol. 1906, no.~1.\hskip 1em plus 0.5em minus 0.4em\relax AIP Publishing LLC,
  2017, p. 120002.

\bibitem{pendleton2017perception}
P.~Scott, A.~Hans, X.~Du, X.~Shen, M.~Malika, E.~You, R.~Daniela, and
  A.~Marcelo, ``Perception, planning, control, and coordination for autonomous
  vehicles,'' \emph{Machines}, vol.~5, no.~1, p.~6, 2017.

\bibitem{paden2016survey}
B.~Paden, M.~{\v{C}}{\'a}p, S.~Z. Yong, D.~Yershov, and E.~Frazzoli, ``A survey
  of motion planning and control techniques for self-driving urban vehicles,''
  \emph{IEEE Transactions on Intelligent Vehicles}, vol.~1, no.~1, pp. 33--55,
  2016.

\bibitem{gonzalez2015review}
D.~Gonz{\'a}lez, J.~P{\'e}rez, V.~Milan{\'e}s, and F.~Nashashibi, ``A review of
  motion planning techniques for automated vehicles,'' \emph{IEEE Transactions
  on Intelligent Transportation Systems}, vol.~17, no.~4, pp. 1135--1145, 2015.

\bibitem{bengler2014three}
K.~Bengler, K.~Dietmayer, B.~Farber, M.~Maurer, and H.~Winner, ``Three decades
  of driver assistance systems: Review and future perspectives,'' \emph{IEEE
  Intelligent Transportation Systems Magazine}, vol.~6, no.~4, pp. 6--22, 2014.

\bibitem{shaout2011advanced}
A.~Shaout, D.~Colella, and S.~Awad, ``Advanced driver assistance systems -
  past, present and future,'' in \emph{2011 Seventh International Computer
  Engineering Conference (ICENCO'2011)}, 2011, pp. 72--82.

\bibitem{hafeez2020insights}
F.~Hafeez, U.~U. Sheikh, N.~Alkhaldi, H.~Z. Al~Garni, Z.~A. Arfeen, and S.~A.
  Khalid, ``Insights and strategies for an autonomous vehicle with a sensor
  fusion innovation: A fictional outlook,'' \emph{IEEE Access}, vol.~8, pp.
  135\,162--135\,175, 2020.

\bibitem{thrun2006stanley}
S.~Thrun, M.~Montemerlo, H.~Dahlkamp, D.~Stavens, A.~Aron, J.~Diebel, P.~Fong,
  J.~Gale, M.~Halpenny, G.~Hoffmann \emph{et~al.}, ``Stanley: The robot that
  won the {DARPA} grand challenge,'' \emph{Journal of Field Robotics}, vol.~23,
  no.~9, pp. p.661--692, 2006.

\bibitem{wang2019pseudo}
Y.~Wang, W.~Chao, D.~Garg, B.~Hariharan, M.~Campbell, and K.~Q. Weinberger,
  ``{Pseudo-LiDAR} from visual depth estimation: Bridging the gap in {3D}
  object detection for autonomous driving,'' in \emph{Proceedings of the
  IEEE/CVF Conference on Computer Vision and Pattern Recognition}, 2019, pp.
  8445--8453.

\bibitem{behroozpour2017lidar}
B.~Behroozpour, P.~A. Sandborn, M.~C. Wu, and B.~E. Boser, ``Lidar system
  architectures and circuits,'' \emph{IEEE Communications Magazine}, vol.~55,
  no.~10, pp. 135--142, 2017.

\bibitem{poulton2017coherent}
C.~V. Poulton, A.~Yaacobi, D.~B. Cole, M.~J. Byrd, M.~Raval, D.~Vermeulen, and
  M.~R. Watts, ``Coherent solid-state {LIDAR} with silicon photonic optical
  phased arrays,'' \emph{Optics Letters}, vol.~42, no.~20, pp. 4091--4094,
  2017.

\bibitem{oniga2009processing}
F.~Oniga and S.~Nedevschi, ``Processing dense stereo data using elevation maps:
  Road surface, traffic isle, and obstacle detection,'' \emph{IEEE Transactions
  on Vehicular Technology}, vol.~59, no.~3, pp. 1172--1182, 2009.

\bibitem{zhang2014surround}
B.~Zhang, V.~Appia, I.~Pekkucuksen, Y.~Liu, A.~Umit~Batur, P.~Shastry, S.~Liu,
  S.~Sivasankaran, and K.~Chitnis, ``A surround view camera solution for
  embedded systems,'' in \emph{Proceedings of the IEEE Conference on Computer
  Vision and Pattern Recognition Workshops}, 2014, pp. 662--667.

\bibitem{appia2015surround}
V.~Appia, H.~Hariyani, S.~Sivasankaran, S.~Liu, K.~Chitnis, M.~Mueller,
  U.~Batur, and G.~Agarwa, ``Surround view camera system for {ADAS on TI¡¯s
  TDAx SoCs},'' \emph{Texas Instruments Technical Note}, 2015.

\bibitem{yoo2020optimization}
M.~W. Yoo and D.~S. Han, ``Optimization algorithm for driver monitoring system
  using deep learning approach,'' in \emph{2020 International Conference on
  Artificial Intelligence in Information and Communication (ICAIIC)}, 2020, pp.
  043--046.

\bibitem{vivacqua2017low}
R.~Vivacqua, R.~Vassallo, and F.~Martins, ``A low cost sensors approach for
  accurate vehicle localization and autonomous driving application,''
  \emph{Sensors}, vol.~17, no.~10, p. 2359, 2017.

\bibitem{patole2017automotive}
S.~M. Patole, M.~Torlak, D.~Wang, and M.~Ali, ``Automotive radars: A review of
  signal processing techniques,'' \emph{IEEE Signal Processing Magazine},
  vol.~34, no.~2, pp. 22--35, 2017.

\bibitem{fan2020computer}
R.~Fan, L.~Wang, M.~J. Bocus, and I.~Pitas, ``Computer stereo vision for
  autonomous driving,'' \emph{arXiv preprint arXiv:2012.03194}, 2020.

\bibitem{preussler2019photonically}
S.~Preussler, F.~Schwartau, J.~Schoebel, and T.~Schneider, ``Photonically
  synchronized large aperture radar for autonomous driving,'' \emph{Optics
  Express}, vol.~27, no.~2, pp. 1199--1207, 2019.

\bibitem{feng2020deep}
D.~Feng, C.~Haase-Sch\"{u}tz, L.~Rosenbaum, H.~Hertlein, C.~Gl\"{a}ser,
  F.~Timm, W.~Wiesbeck, and K.~Dietmayer, ``Deep multi-modal object detection
  and semantic segmentation for autonomous driving: Datasets, methods, and
  challenges,'' \emph{IEEE Transactions on Intelligent Transportation Systems},
  pp. 1--20, 2020.

\bibitem{aydogdu2020radar}
C.~Aydogdu, M.~F. Keskin, G.~K. Carvajal, O.~Eriksson, H.~Hellsten,
  H.~Herbertsson, E.~Nilsson, M.~Rydstrom, K.~Vanas, and H.~Wymeersch, ``Radar
  interference mitigation for automated driving: Exploring proactive
  strategies,'' \emph{IEEE Signal Processing Magazine}, vol.~37, no.~4, pp.
  72--84, 2020.

\bibitem{ingle2016tesla}
S.~Ingle and M.~Phute, ``Tesla autopilot: Semi autonomous driving, an uptick
  for future autonomy,'' \emph{International Research Journal of Engineering
  and Technology}, vol.~3, no.~9, pp. 369--372, 2016.

\bibitem{alonso2011ultrasonic}
L.~Alonso, V.~Milan\'{e}s, C.~Torre-Ferrero, J.~Godoy, J.~P. Oria, and
  T.~De~Pedro, ``Ultrasonic sensors in urban traffic driving-aid systems,''
  \emph{Sensors}, vol.~11, no.~1, pp. 661--673, 2011.

\bibitem{rhee2019low}
J.~H. Rhee and J.~Seo, ``Low-cost curb detection and localization system using
  multiple ultrasonic sensors,'' \emph{Sensors}, vol.~19, no.~6, p. 1389, 2019.

\bibitem{luo2019localization}
Q.~Luo, Y.~Cao, J.~Liu, and A.~Benslimane, ``Localization and navigation in
  autonomous driving: Threats and countermeasures,'' \emph{IEEE Wireless
  Communications}, vol.~26, no.~4, pp. 38--45, 2019.

\bibitem{kastrinaki2003survey}
V.~Kastrinaki, M.~Zervakis, and K.~Kalaitzakis, ``A survey of video processing
  techniques for traffic applications,'' \emph{Image and Vision Computing},
  vol.~21, no.~4, pp. 359--381, 2003.

\bibitem{liu2017computer}
S.~Liu, J.~Tang, Z.~Zhang, and J.-L. Gaudiot, ``Computer architectures for
  autonomous driving,'' \emph{Computer}, vol.~50, no.~8, pp. 18--25, 2017.

\bibitem{campbell2018sensor}
S.~Campbell, N.~O'Mahony, L.~Krpalcova, D.~Riordan, J.~Walsh, A.~Murphy, and
  C.~Ryan, ``Sensor technology in autonomous vehicles: A review,'' in
  \emph{2018 29th Irish Signals and Systems Conference (ISSC)}.\hskip 1em plus
  0.5em minus 0.4em\relax IEEE, 2018, pp. 1--4.

\bibitem{zekavat2011handbook}
R.~Zekavat and R.~M. Buehrer, \emph{Handbook of Position Location: Theory,
  Practice and Advances}.\hskip 1em plus 0.5em minus 0.4em\relax John Wiley \&
  Sons, 2011, vol.~27.

\bibitem{mogelmose2012Vision}
A.~Mogelmose, M.~M. Trivedi, and T.~B. Moeslund, ``Vision-based traffic sign
  detection and analysis for intelligent driver assistance systems:
  Perspectives and survey,'' \emph{IEEE Transactions on Intelligent
  Transportation Systems}, vol.~13, no.~4, pp. 1484--1497, 2012.

\bibitem{he2019traffic}
Z.~He, F.~Nan, X.~Li, S.-J. Lee, and Y.~Yang, ``Traffic sign recognition by
  combining global and local features based on semi-supervised
  classification,'' \emph{IET Intelligent Transport Systems}, vol.~14, no.~5,
  pp. 323--330, 2019.

\bibitem{arcos2018evaluation}
A.~Arcos-Garc\'{\i}a, J.~A. \'{A}lvarez Garc\'{\i}a, and L.~M. Soria-Morillo,
  ``Evaluation of deep neural networks for traffic sign detection systems,''
  \emph{Neurocomputing}, vol. 316, pp. 332--344, 2018.

\bibitem{arcos2018deep}
{\'A}.~Arcos-Garc{\'\i}a, J.~A. \'{A}lvarez Garc\'{\i}a, and L.~M.
  Soria-Morillo, ``Deep neural network for traffic sign recognition systems: An
  analysis of spatial transformers and stochastic optimisation methods,''
  \emph{Neural Networks}, vol.~99, pp. 158--165, 2018.

\bibitem{liang2019traffic}
Z.~Liang, J.~Shao, D.~Zhang, and L.~Gao, ``Traffic sign detection and
  recognition based on pyramidal convolutional networks,'' \emph{Neural
  Computing and Applications}, pp. 1--11, 2019.

\bibitem{zhang2020cascaded}
J.~Zhang, Z.~Xie, J.~Sun, X.~Zou, and J.~Wang, ``A cascaded {R-CNN} with
  multiscale attention and imbalanced samples for traffic sign detection,''
  \emph{IEEE Access}, vol.~8, pp. 29\,742--29\,754, 2020.

\bibitem{liu2020small}
Z.~Liu, D.~Li, S.~S. Ge, and F.~Tian, ``Small traffic sign detection from large
  image,'' \emph{Applied Intelligence}, vol.~50, no.~1, pp. 1--13, 2020.

\bibitem{guan2018robust}
H.~Guan, W.~Yan, Y.~Yu, L.~Zhong, and D.~Li, ``Robust traffic-sign detection
  and classification using mobile {LiDAR} data with digital images,''
  \emph{IEEE Journal of Selected Topics in Applied Earth Observations and
  Remote Sensing}, vol.~11, no.~5, pp. 1715--1724, 2018.

\bibitem{guan2019convolutional}
H.~Guan, Y.~Yu, D.~Peng, Y.~Zang, J.~Lu, A.~Li, and J.~Li, ``A convolutional
  capsule network for traffic-sign recognition using mobile {LiDAR} data with
  digital images,'' \emph{IEEE Geoscience and Remote Sensing Letters}, vol.~17,
  no.~6, pp. 1067--1071, 2019.

\bibitem{balado2020novel}
J.~Balado, E.~Gonz{\'a}lez, P.~Arias, and D.~Castro, ``Novel approach to
  automatic traffic sign inventory based on mobile mapping system data and deep
  learning,'' \emph{Remote Sensing}, vol.~12, no.~3, p. 442, 2020.

\bibitem{hirabayashi2019traffic}
M.~Hirabayashi, A.~Sujiwo, A.~Monrroy, S.~Kato, and M.~Edahiro, ``Traffic light
  recognition using high-definition map features,'' \emph{Robotics and
  Autonomous Systems}, vol. 111, pp. 62--72, 2019.

\bibitem{mannan2019classification}
A.~Mannan, K.~Javed, A.~Ur-Rehman, H.~A. Babri, and S.~K. Noon,
  ``Classification of degraded traffic signs using flexible mixture model and
  transfer learning,'' \emph{IEEE Access}, vol.~7, pp. 148\,800--148\,813,
  2019.

\bibitem{zhang2020lightweight}
J.~Zhang, W.~Wang, C.~Lu, J.~Wang, and A.~K. Sangaiah, ``Lightweight deep
  network for traffic sign classification,'' \emph{Annals of
  Telecommunications}, vol.~75, no.~7, pp. 369--379, 2020.

\bibitem{xing2018advances}
X.~Yang, L.~Chen, L.~Chen, H.~Wang, H.~Wang, D.~Cao, E.~Velenis, and F.~Wang,
  ``Advances in vision-based lane detection: Algorithms, integration,
  assessment, and perspectives on {ACP}-based parallel vision,'' \emph{IEEE/CAA
  Journal of Automatica Sinica}, vol.~5, no.~3, pp. 645--661, 2018.

\bibitem{chen2019deep}
L.~Chen, W.~Zhan, W.~Tian, Y.~He, and Q.~Zou, ``Deep integration: A multi-label
  architecture for road scene recognition,'' \emph{IEEE Transactions on Image
  Processing}, vol.~28, no.~10, pp. 4883--4898, 2019.

\bibitem{bayoudh2020transfer}
K.~Bayoudh, F.~Hamdaoui, and A.~Mtibaa, ``Transfer learning based hybrid {2D-3D
  CNN} for traffic sign recognition and semantic road detection applied in
  advanced driver assistance systems,'' \emph{Applied Intelligence}, pp. 1--19,
  2020.

\bibitem{guindel2019traffic}
C.~Guindel, D.~Martin, and J.~M. Armingol, ``Traffic scene awareness for
  intelligent vehicles using {ConvNets} and stereo vision,'' \emph{Robotics and
  Autonomous Systems}, vol. 112, pp. 109--122, 2019.

\bibitem{ye2018lane}
Y.~Ye, X.~Hao, and H.~Chen, ``Lane detection method based on lane structural
  analysis and {CNNs},'' \emph{IET Intelligent Transport Systems}, vol.~12,
  no.~6, pp. 513--520, 2018.

\bibitem{zou2019robust}
Q.~Zou, H.~Jiang, Q.~Dai, Y.~Yue, L.~Chen, and Q.~Wang, ``Robust lane detection
  from continuous driving scenes using deep neural networks,'' \emph{IEEE
  Transactions on Vehicular Technology}, vol.~69, no.~1, pp. 41--54, 2019.

\bibitem{li2016deep}
J.~Li, X.~Mei, D.~Prokhorov, and D.~Tao, ``Deep neural network for structural
  prediction and lane detection in traffic scene,'' \emph{IEEE Transactions on
  Neural Networks and Learning Systems}, vol.~28, no.~3, pp. 690--703, 2016.

\bibitem{hou2019learning}
Y.~Hou, Z.~Ma, C.~Liu, and C.~C. Loy, ``Learning lightweight lane detection
  {CNNs} by self attention distillation,'' in \emph{Proceedings of the IEEE/CVF
  International Conference on Computer Vision}, 2019, pp. 1013--1021.

\bibitem{xiao2020attention}
D.~Xiao, X.~Yang, J.~Li, and M.~Islam, ``Attention deep neural network for lane
  marking detection,'' \emph{Knowledge-Based Systems}, vol. 194, p. 105584,
  2020.

\bibitem{song2018lane}
W.~Song, Y.~Yang, M.~Fu, Y.~Li, and M.~Wang, ``Lane detection and
  classification for forward collision warning system based on stereo vision,''
  \emph{IEEE Sensors Journal}, vol.~18, no.~12, pp. 5151--5163, 2018.

\bibitem{xiao2018hybrid}
L.~Xiao, R.~Wang, B.~Dai, Y.~Fang, D.~Liu, and T.~Wu, ``Hybrid conditional
  random field based camera-{LIDAR} fusion for road detection,''
  \emph{Information Sciences}, vol. 432, pp. 543--558, 2018.

\bibitem{caltagirone2019lidar}
L.~Caltagirone, M.~Bellone, L.~Svensson, and M.~Wahde, ``Lidar-camera fusion
  for road detection using fully convolutional neural networks,''
  \emph{Robotics and Autonomous Systems}, vol. 111, pp. 125--131, 2019.

\bibitem{chen2019progressive}
Z.~Chen, J.~Zhang, and D.~Tao, ``Progressive {LiDAR} adaptation for road
  detection,'' \emph{IEEE/CAA Journal of Automatica Sinica}, vol.~6, no.~3, pp.
  693--702, 2019.

\bibitem{kim2018pedestrian}
J.~H. Kim, G.~Batchuluun, and K.~R. Park, ``Pedestrian detection based on
  faster {R-CNN} in nighttime by fusing deep convolutional features of
  successive images,'' \emph{Expert Systems with Applications}, vol. 114, pp.
  15--33, 2018.

\bibitem{li2019deep}
G.~Li, Y.~Yang, and X.~Qu, ``Deep learning approaches on pedestrian detection
  in hazy weather,'' \emph{IEEE Transactions on Industrial Electronics},
  vol.~67, no.~10, pp. 8889--8899, 2019.

\bibitem{guan2019fusion}
D.~Guan, Y.~Cao, J.~Yang, Y.~Cao, and M.~Y. Yang, ``Fusion of multispectral
  data through illumination-aware deep neural networks for pedestrian
  detection,'' \emph{Information Fusion}, vol.~50, pp. 148--157, 2019.

\bibitem{henein2020dynamic}
M.~Henein, J.~Zhang, R.~Mahony, and V.~Ila, ``Dynamic {SLAM}: The need for
  speed,'' in \emph{2020 IEEE International Conference on Robotics and
  Automation (ICRA)}, 2020, pp. 2123--2129.

\bibitem{wang2017pedestrian}
H.~Wang, B.~Wang, B.~Liu, X.~Meng, and G.~Yang, ``Pedestrian recognition and
  tracking using {3D LiDAR} for autonomous vehicle,'' \emph{Robotics and
  Autonomous Systems}, vol.~88, pp. 71--78, 2017.

\bibitem{ali2018yolo3d}
W.~Ali, S.~Abdelkarim, M.~Zahran, M.~Zidan, and A.~E. Sallab, ``{YOLO3D}:
  End-to-end real-time {3D} oriented object bounding box detection from {LiDAR}
  point cloud,'' in \emph{Proceedings of the European Conference on Computer
  Vision (ECCV) Workshops}, 2018, pp. 0--0.

\bibitem{zhou2018voxelnet}
Y.~Zhou and O.~Tuzel, ``{VoxelNet}: End-to-end learning for point cloud based
  {3D} object detection,'' in \emph{Proceedings of the IEEE Conference on
  Computer Vision and Pattern Recognition}, 2018, pp. 4490--4499.

\bibitem{yang2020multi}
Y.~Yang, F.~Chen, F.~Wu, D.~Zeng, Y.~Ji, and X.~Jing, ``Multi-view semantic
  learning network for point cloud based {3D} object detection,''
  \emph{Neurocomputing}, vol. 397, pp. 477--485, 2020.

\bibitem{asvadi2018multimodal}
A.~Asvadi, L.~Garrote, C.~Premebida, P.~Peixoto, and U.~J.~Nunes, ``Multimodal
  vehicle detection: fusing {3D-LIDAR} and color camera data,'' \emph{Pattern
  Recognition Letters}, vol. 115, pp. 20--29, 2018.

\bibitem{gao2018object}
H.~Gao, B.~Cheng, J.~Wang, K.~Li, J.~Zhao, and D.~Li, ``Object classification
  using {CNN}-based fusion of vision and {LIDAR} in autonomous vehicle
  environment,'' \emph{IEEE Transactions on Industrial Informatics}, vol.~14,
  no.~9, pp. 4224--4231, 2018.

\bibitem{zhao2020fusion}
X.~Zhao, P.~Sun, Z.~Xu, H.~Min, and H.~Yu, ``Fusion of {3D LIDAR} and camera
  data for object detection in autonomous vehicle applications,'' \emph{IEEE
  Sensors Journal}, vol.~20, no.~9, pp. 4901--4913, 2020.

\bibitem{chen20173d}
X.~Chen, K.~Kundu, Y.~Zhu, H.~Ma, S.~Fidler, and R.~Urtasun, ``{3D} object
  proposals using stereo imagery for accurate object class detection.''
  \emph{IEEE Transactions on Pattern Analysis and Machine Intelligence},
  vol.~40, no.~5, pp. 1259--1272, 2017.

\end{thebibliography}



\end{document}